# Driving Intelligent IoT Monitoring and Control through Cloud Computing and Machine Learning

**Hanzhe Li** [1]*

Computer Engineering,New York University,NY USA

* Corresponding author:Nyhanzheli@gmail.com

**Xiangxiang Wang** [1]

Computer Science,University of Texas at Arlington,Arlington, TX,USA

wx18714999@gmail.com

**Yuan Feng** [2]

Interdisciplinary Data Science ,Duke University ,North Carolina USA

yuan.feng.dsduke@gmail.com

**Yaqian Qi** [3]

Quantitative Methods and Modeling,Baruch Collegue, CUNY ,55 Lexington Ave, NY,USA

alicia.qi.yaqian@gmail.com

**Jingxiao Tian** [4]

Electrical and Computer Engineering,San Diego State University,SD, USA

jtian1125@sdsu.edu

**Abstract**

This article explores how to drive intelligent iot monitoring and control through cloud computing and machine learning. As iot and the cloud continue to generate large and diverse amounts of data as sensor devices in the network, the collected data is sent to the cloud for statistical analysis, prediction, and data analysis to achieve business objectives. However, because the cloud computing model is limited by distance, it can be problematic in environments where the quality of the Internet connection is not ideal for critical operations. Therefore, edge computing, as a distributed computing architecture, moves the location of processing applications, data and services from the central node of the network to the logical edge node of the network to reduce the dependence on cloud processing and analysis of data, and achieve near-end data processing and analysis. The combination of iot and edge computing can reduce latency, improve efficiency, and enhance security, thereby driving the development of intelligent systems. The paper also introduces the development of iot monitoring and control technology, the application of edge computing in iot monitoring and control, and the role of machine learning in data analysis and fault detection. Finally, the application and effect of intelligent Internet of Things monitoring and control system in industry, agriculture, medical and other fields are demonstrated through practical cases and experimental studies.

# 1 INTRODUCTION

The origins of iot monitoring and control technology can be traced back to the 1980s, when remote monitoring and control systems began to appear in the field of industrial automation. With the development of computer and communication technology, people began to try to connect sensors and actuators with the network to achieve remote monitoring and control. In 1999, an MIT study first proposed the concept of the "Internet of Things," meaning that objects can be connected and communicated with each other through a network, enabling information sharing and intelligent control. Since then, Internet of Things monitoring and control technology has gradually become a research hotspot in industry, agriculture, health and other fields. Key technological breakthroughs include the development of sensor technology, advances in wireless communication technology, and improvements in data processing and analysis algorithms. With the continuous evolution of Internet of Things technology, more and more application scenarios have been realized, such as smart homes, smart cities, smart factories, etc., bringing convenience and efficiency improvement to people's life and work.

With the gradual maturity of Internet of Things technology, Internet of Things monitoring and control have been widely used in various fields. In the industrial sector, iot monitoring and control technology can enable intelligent manufacturing, including equipment condition monitoring, production process optimization and predictive maintenance. In agriculture, iot monitoring and control technology can enable precision agriculture, including soil moisture monitoring, crop growth forecasting, and smart irrigation. In the field of health, iot monitoring and control technology can enable remote health monitoring, including heart rate monitoring, sleep quality assessment and disease prevention. In addition, the Internet of Things monitoring and control technology can also be applied to environmental monitoring, traffic management, energy management and other aspects, providing important support for the sustainable development of the social economy.

In recent years, with the continuous maturity of Internet of Things technology and driven by relevant national policies, a large number of innovative applications in the Internet of Things industry have been rapidly developed. From the explosive growth of consumer smart homes and smart items to the continuous innovation of enterprise in intelligent manufacturing, intelligent transportation, public safety and medical fields, the market size of the entire Internet of Things is expanding rapidly. However, the ensuing problem is that more and more Internet of Things information security incidents are frequent.

## 2 RELATED WORK

### 2.1 Application of edge computing in iot monitoring and control

Edge computing (also known as Edge computing) is a distributed computing architecture that moves the processing of applications, data, and services from the central node of the network to the logical edge nodes of the network. Edge computing will be completely handled by the central node to decompose large services, cut into smaller and more manageable parts, scattered to the edge node to deal with. The edge node is closer to the user terminal device, which can speed up the processing and transmission of data and reduce the delay. In this framework, the analysis of data and the generation of knowledge are closer to the source of data, so it is more suitable for dealing with big data.

The "edge" is defined as any compute and network resource along the path between the data source and the cloud data center. For example, the smartphone is the edge between the user and the cloud, the gateway in the smart home is the edge between the home and the cloud, and micro data centers and cloudlets are the edge between mobile devices and the cloud. The basic principle of edge computing is that the computation should occur in close proximity to the data source for processing. From a researcher's perspective, edge computing and fog computing are interchangeable, but edge computing is more focused on the thing side, while fog computing is more focused on the infrastructure side.

Data is increasingly found at the edge of the network, where it can be processed more efficiently. Previous work such as micro datacenter, cloudlet and fog computing has been introduced into the community because cloud computing is not always effective for data processing when data is generated at the edge of the network. In many cases, edge computing is more efficient than cloud computing for some computing services.

Edge computing can effectively solve the connectivity problems faced by iot devices. By moving critical data processing functions to the edge of the network or near the source of the data, edge computing can help connected devices maintain the same level of efficiency even when the network connection is poor.

## 2.2 Application Research and Practical Cases

In the dynamic landscape of IoT monitoring and control, the convergence of machine learning techniques has emerged as a catalyst for transformative innovation. Here, we explore the dual facets of application research and practical cases, showcasing the symbiotic relationship between machine learning and intelligent IoT systems:

### (1) Data Analysis and Predictive Maintenance

Machine learning algorithms empower IoT systems to extract actionable insights from vast datasets, enabling advanced data analysis. Key techniques include:

$$\sum i = 1k \sum x \in S_i \; ||x - ui||_2^2 \quad (1)$$

Anomaly Detection: $p(x)<\epsilon$

Research demonstrates the efficacy of these algorithms in identifying patterns and anomalies, optimizing system performance, and facilitating predictive maintenance strategies.

### (2) Energy Optimization and Fault Detection:

Machine learning enhances energy optimization efforts by analyzing consumption patterns and dynamically adjusting resource allocation in IoT-enabled systems. Key formulas include:

$$\tilde{y}=f(x) \quad (2)$$

(3) Resource Allocation:

$$\sum_{i=1}^{n} \omega_i x_i \quad (3)$$

Real-time fault detection and diagnosis benefit from machine learning algorithms, ensuring system reliability and uptime through prompt remedial actions.

By integrating machine learning into intelligent IoT monitoring and control systems, industries unlock unprecedented levels of efficiency, productivity, and sustainability, shaping a future where technology seamlessly integrates with everyday life.

## 3  APPLICATION RESEARCH AND PRACTICAL CASES

### 3.1  Data Analysis and Predictive Maintenance

Machine learning can make real-time predictions. Through the Internet of Things and clustering algorithms, we can monitor and optimize production processes to ensure the safety of workers in dangerous areas. For example, consider the production process of a chemical plant. Data collected using iot sensors, such as temperature, pressure, and chemical concentrations, can be fed into a clustering algorithm for analysis. By clustering historical and current data, potential production risks can be predicted and timely measures can be taken to avoid accidents. In chemical plants, for example, clustering algorithms can identify different production states and possible anomalies based on past data and current real-time data. Once an abnormal state is found, the system can automatically take measures, such as adjusting production parameters or issuing alarms, to ensure the safety and stability of the production process.

In addition, iot analytics can also help save costs in industrial applications. The traditional concept of "scheduled maintenance" will become obsolete, and we can now look forward to using predictive maintenance to reduce unplanned downtime. Predictive maintenance is a preventive measure designed to prevent machine failures from affecting production. But maintenance itself also stops production and may require checking all the machines on the entire line. By analyzing iot data through clustering algorithms, we can identify potential failure patterns and take steps to prevent downtime events. Modern machines utilize iot sensors to monitor a variety of data, including usage, uptime, energy consumption, and system failure logs. Historical data analysis and predictive analytics through machine learning models can help us understand the life cycle of parts and identify production quality degradation due to faulty parts. The combination of iot and clustering algorithms helps to achieve effective risk management. Clustering algorithms can use past data to predict a risk and automatically react to that risk.

## 3.2 3.2 Introduction to Experimental Framework: IoT and Clustering Algorithms

To validate the effectiveness of machine learning algorithms in IoT environments, experimental frameworks are essential. In this section, we introduce the experimental setup focusing on IoT and clustering algorithms. Machine learning relies on data, which spans various types, including numerical, visual, or textual data such as sensor data, images, or textual records. This section outlines the data collection and preparation process for training machine learning models in our experimental setup.These two measures are similar in principle. DI is chosen as the measurement index for analyzing clustering performance in this paper. Assume that the:

$$DD= \{1, 2, \ldots x_m, \} \quad (1)$$

The cluster is defined as follows:

$$DI = \min_{1\leq i\leq k}\{\min_{j\neq i}(\frac{d_{min}(C_i, C_j)}{\max_{1\leq l\leq k} diam\ (C_l)})\} \quad (2)$$

The denominator in the above equation corresponds to the maximum distance between the target points in all clusters, and the numerator corresponds to the minimum distance between the target points in all clusters. Obviously, for the same data set, the larger the DI corresponding to a cluster result, the better the performance of the cluster. That is, the smaller the maximum distance, the more concentrated the point cloud, the more similar the point cloud.

## 3.3 Energy Optimization and Fault Detection

In the Internet of Things (IoT) environment, real-time fault detection and diagnosis is a key component to ensuring system reliability and continuous operation. By combining machine learning algorithms and real-time data monitoring, the system is able to react quickly when failures occur and take timely corrective action to minimize downtime and repair costs, ensuring continuous and stable operation of the system. Real-time fault detection relies on real-time data from monitoring iot devices and sensors. This data includes equipment status, sensor readings, environmental conditions, and more. Machine learning algorithms are applied to this data to identify abnormal patterns and potential signals of failure. Common machine learning techniques include supervised learning, unsupervised learning, and deep learning. Supervised learning is used to build a model of the normal working mode of the device to detect anomalies. Unsupervised learning is used to find abnormal patterns in data, while deep learning is used to process complex, high-dimensional data and extract features to identify potential fault signals.

Once an exception has been detected, the next step is troubleshooting. This involves determining the specific cause of the failure so that appropriate corrective action can be taken. Troubleshooting typically involves the following key steps:

1. Feature extraction and selection: Fault related features are extracted from the monitored data. This may involve techniques such as data dimensionality reduction, feature selection, and feature engineering to improve diagnostic accuracy and efficiency.

2. Pattern recognition and classification: Machine learning algorithms are used to analyze and classify the extracted features to identify different types of failure modes. Common methods include support vector machines, decision trees and neural networks.

3. Fault location and reasoning: Based on the identified fault mode, infer the possible fault location and cause. This may involve the intellectual reasoning of expert systems, or the use of probabilistic models to assess the likelihood of different causes of failure.

4. Corrective actions and feedback: Take appropriate corrective actions based on the diagnosis result, such as repairing the device, replacing parts, or adjusting the system configuration. At the same time, the diagnosis results are fed back to the system operations personnel so that they can understand the current system status and take necessary actions.

### 3.4 Practical Cases across Industries

Intelligent IoT monitoring and control systems have revolutionized operations across various industries, offering unprecedented levels of efficiency, productivity, and insights. By harnessing the power of IoT devices, coupled with advanced analytics and machine learning algorithms, businesses can optimize their processes, improve decision-making, and enhance overall performance. Let's explore some practical applications across different sectors:

1. Industry 4.0: Predictive Maintenance and Real-Time Equipment Monitoring

In the realm of Industry 4.0, intelligent IoT monitoring and control systems play a pivotal role in predictive maintenance and real-time equipment monitoring. By deploying sensors and connected devices throughout the production line, manufacturers can gather real-time data on equipment performance, detect anomalies, and predict potential failures before they occur. This proactive approach helps minimize downtime, reduce maintenance costs, and maximize operational efficiency, ultimately driving greater competitiveness in the market.

2. Smart Agriculture: Soil Moisture Monitoring and Crop Yield Prediction

In agriculture, intelligent IoT systems enable farmers to monitor crucial environmental factors such as soil moisture levels, temperature, and humidity in real time. By leveraging this data, farmers can make informed decisions about irrigation scheduling, fertilizer application, and crop management practices. Additionally, machine learning algorithms can analyze historical data and environmental patterns to predict crop yields with remarkable accuracy. This empowers farmers to optimize resource allocation, maximize harvests, and ensure sustainable agricultural practices.

3. Healthcare: Remote Patient Monitoring and Disease Prediction

The healthcare industry benefits immensely from intelligent IoT monitoring and control systems, particularly in the realm of remote patient monitoring and disease prediction. Wearable devices and medical sensors enable continuous monitoring of vital signs, medication adherence, and other health metrics in real time, allowing healthcare providers to remotely track patients' health status and intervene promptly in case of any abnormalities. Moreover, by analyzing vast amounts of patient data, machine learning algorithms can identify patterns and trends associated with various diseases, enabling early detection, intervention, and personalized treatment plans.

In summary, intelligent IoT monitoring and control systems are driving transformative changes across industries, revolutionizing traditional practices and unlocking new opportunities for innovation and growth. By harnessing the power of connected devices, advanced analytics, and machine learning, businesses can achieve unprecedented levels of efficiency, productivity, and competitiveness in today's dynamic market landscape.

## 4 EXPERIMENT METHODOLOGY AND RESULTS

### 4.1 IoTEnsemble: Machine learning-based iot attack detection

In this paper, IoTEnsemble, a machine learning-based network anomaly detection framework, lays the core foundation of this paper. Firstly, an activity clustering method based on tree structure is proposed, which can effectively identify and aggregate network flows for the same device activity. Based on the clustering results, it can profile the traffic pattern of each device behavior separately, thus effectively reducing the dependence on the generalization of a single model.

In the experimental phase, the researchers examined a 57.1 gigabyte iot traffic dataset spanning nine months and a rich malicious traffic dataset. The results show that IoTEnsemble can detect multiple botnet malware and cyberattacks more effectively than existing solutions in today's intelligent and functional iot environments.

### 4.2 Traffic activities implement clustering

The ultimate goal of the activity clustering method proposed in this paper is to aggregate the traffic used for the same activity so that each sub-model in the integrated model learns the traffic model of a single activity. The algorithm needs to meet three design goals: 1) good interpretability, which can be understood by network administrators; 2) as little prior knowledge as possible, such as the number of protocols or activities used by the device; 3) Appropriate clustering granularity to ensure the correctness of clustering while using as few clusters as possible.

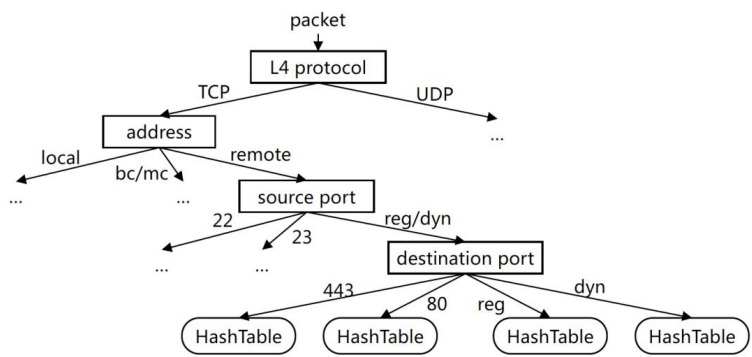

**Figure 1:** Activity clustering based on tree structure; bc/mc

An Internet of Things behavior clustering method based on tree structure is proposed (as shown in Figure 1). First, a packet from the bidirectional stream quintuple f=(device-IP, dst-domain/IP, src-port, dst-port, protocol) is initially clustered using four-level rules:

- L4 protocol: TCP or UDP;
- Address: dst-domain/IP is a resolved domain name, remote IP address, local IP address, or broadcast/multicast address;
- src-port: a system port number or a registered port or dynamic port number range;
- dst-port: indicates a system port number, registered port number range or dynamic port number range.

Since the iot device is generally on the client side, the source port number is often not helpful for clustering (unless it is a system port such as 22, 23 for SSH, Telnet), but the destination port in the registered port number range can still represent some common iot services, such as SSDP (1900), STUN (3478), etc.

At the end of the tree structure, each leaf node contains a hash table, whose key value IS a bidirectional stream quintuple f, whose value is an incremental statistical structure IS=(Nin, Nout, Tin, Tout, S), where N represents the number of packets, T represents the sum of packet arrival intervals, in and out represent the direction, and so on. S represents the set of package sizes. No matter how many packets a stream has, this structure can be maintained at a constant level of storage complexity. When a packet with direction d, packet size s, and arrival interval t enters a leaf node of the tree structure, IS updates it in the following way:

$$
\begin{aligned}
&N^{in} \leftarrow N^{in} + 1 \text{ if } d = in \text{ else } N^{out} \leftarrow N^{out} + 1 \\
&T^{in} \leftarrow T^{in} + \Delta t \text{ if } d = in \text{ else } T^{out} \leftarrow T^{out} + \Delta t \\
&\mathbb{S}.add(s)
\end{aligned}
\tag{3}
$$

More abstract flow rules can be obtained by merging internal quintuples that belong to the same activity. For the two quintuples of f1 and f2, their statistics IS1 and IS2 are compared to determine whether they belong to the same activity: 1. Spatial correlation: Jaccard index is used to compare two packet size sets S1 and S2. If they exceed a certain threshold hs, they are considered to have spatial correlation.

### 4.3 Model construction

IoTEnsemble has two detection stages:

1. Rule matching: Using active key values can quickly filter out highly abnormal traffic, such as connections with unknown domain names and unknown protocols;

2. Integrated model: Learn traffic patterns for each activity and detect abnormal traffic by identifying deviations from normal patterns.

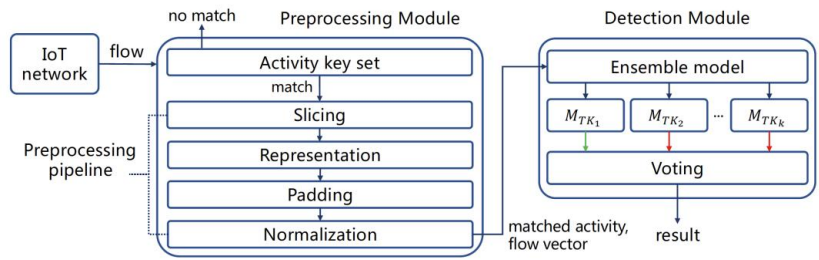

Figure 2: IoTEnsemble overall architecture

Figure 2 shows the overall structure of IoTEnsemble. The pre–processing module first receives iot traffic and uses a quintuple to match the set of active key values obtained through the clustering tree. Since the clustering tree algorithm results in an abstract plan (such as domain names containing wildcards, "reg/dyn" port number ranges, etc.), the matching is a fuzzy matching process, which reduces the possibility of false positives. But if the match fails, the stream is directly marked as malicious.

Traffic matched by active key values enters the second detection phase based on ensemble learning. To implement machine learning–based processing, the preprocessing pipeline gets a numerical representation of the data from a stream's packets. The pipeline includes four steps: segmentation, characterization, filling and normalization, and finally a sequence feature vector is constructed using the IP packet length and arrival time interval of the first r packets of a quintuple.

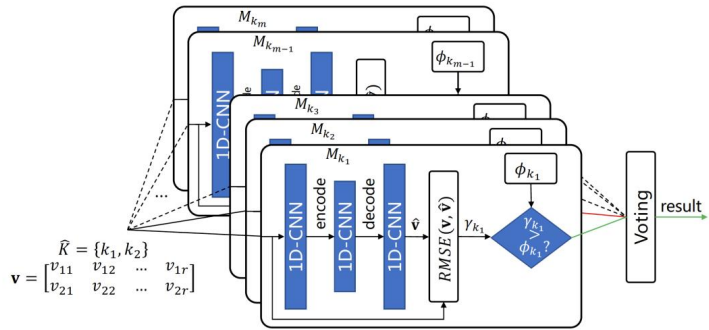

Figure 3. Detection module based on ensemble learning

The detection module is an integrated model composed of multiple unsupervised learning submodels, each of which learns an activity's traffic pattern independently (as shown in Figure 3). The advantages of unsupervised learning include the absence of malicious traffic for training and the ability to detect zero-day attacks. Each submodel uses a 1D convolutional neural network autoencoder (1D-CNN AE), which can learn the implicit representation of the data through a compression-decompression process. In the detection phase, data that does not conform to the learning distribution generates a higher reconstruction error and is thus detected. In addition, in order to reduce the running overhead, the model uses a trigger-action mechanism. Specifically, a stream only needs to trigger the detection of the submodels corresponding to those active key values that are fuzzy matched, and other models do not need to be awakened.

### 4.4 Verification result

Six existing NADS were used as baselines to compare the abnormal traffic detection capabilities of the IoTEnsemble framework, including Kitsune, the SOTA solution published in NDSS. As can be seen in Figure 4, IoTEnsemble achieves better detection effect than other schemes for most attack categories. We believe that this effect is mainly due to the fact that effective and reliable activity clustering can aggregate traffic data that is more consistent with i.i.d., thus significantly reducing the learning difficulty of each submodel. In contrast, Kitsune's performance on some of the more complex devices, such as cameras and audio, became significantly worse, while IoTEnsemble's detection ability was barely affected.

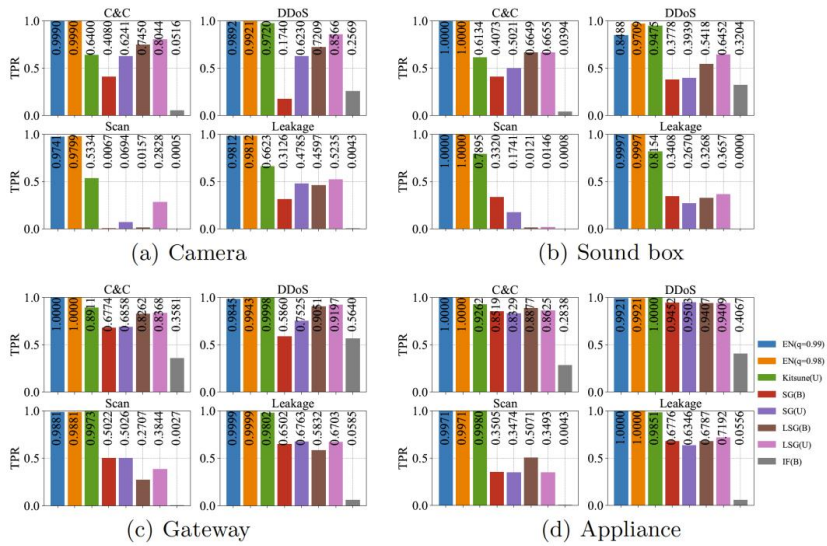

Figure 5. Comparison of IoTEnsemble and five baseline NADS

It can be seen from the results in FIG. 5 that in machine learn-based iot fault detection, malware can evolve to use common protocols such as HTTP to disguise its C&C channel. We encapsulate the C&C of four types of malware using HTTP headers and send malicious traffic through port 80. This makes it possible for malicious traffic to avoid IoTEnsemble's first detection phase based on rule matching. Even so, the experimental results show that IoTEnsemble can still detect these covert attack traffic with a high probability.

Therefore, according to IoTEnsemble, an anomaly detection framework for iot botnets and related network attacks, it can be seen that the framework is designed to adapt to the current increasingly powerful new iot devices. It has two detection stages, including an active clustering algorithm based on tree structure and an integration model. We have built a real iot experimental platform that fully demonstrates device functionality. The experimental results show that IoTEnsemble achieves leading detection against multiple attacks regardless of the amount of activity in the iot network.

## 5 CONCLUSION

To sum up, this experiment adopts advanced machine learning algorithm and network security technology, designs an anomaly detection system based on IoTEnsemble framework, and verifies it in a real IoT environment. The experimental results show that IoTEnsemble has significant performance advantages in detecting a variety of network attacks and malicious activities, showing higher detection accuracy and robustness compared to existing schemes. This achievement is not only of great significance for improving the level of IoT network security, but also provides useful reference and reference for the research and practice in related fields. However, this experiment also has some limitations, such as the size and diversity of the dataset, which need to be further improved. Future research directions can include expanding experiment scale, optimizing algorithm performance, and exploring IoT network security issues to further improve and improve the performance and applicability of the anomaly detection system.

This paper explores the methods and technologies that drive intelligent iot monitoring and control through cloud computing and machine learning. From the point of view that iot and cloud computing continue to generate large amounts of data, this paper discusses the application of edge computing as a distributed computing architecture, and the role of machine learning in data analysis and fault detection. By combining iot with edge computing, latency can be reduced, efficiency improved, and security enhanced, thereby driving the development of intelligent systems. The experimental part shows the application of machine learning algorithm in anomaly detection in iot environment, which provides strong support for building a safer and more reliable iot environment. In the future, we look forward to further exploring and applying new technologies and methods, constantly promoting the innovation and development of iot monitoring and control technology, and making more contributions to building intelligent, efficient and secure iot systems."